\documentclass[11pt]{article}
\usepackage{amsmath,amssymb}
\usepackage{graphicx} 
\usepackage{url}
\usepackage[colorlinks=true, allcolors=blue]{hyperref} 
\usepackage[linesnumbered,ruled,vlined]{algorithm2e}
\usepackage[margin=1in]{geometry}
\newenvironment{keywords}{\noindent\textbf{Keywords:} }{}

\title{A Robust PPO-optimized Tabular Transformer Framework for Intrusion Detection in Industrial IoT Systems}
\author{
\quad Russell. Y. She$^{*,1}$ \qquad \\
\small $^1$Weixi Technology, Hangzhou, China \\
\small $^*$Corresponding author: \texttt{russellshe@gmail.com}
}
\date{May 2025}

\begin{document}

\maketitle

\begin{abstract}
In this paper, we propose a robust and reinforcement-learning-enhanced network intrusion detection system (NIDS) designed for class-imbalanced and few-shot attack scenarios in Industrial Internet of Things (IIoT) environments. Our model integrates a TabTransformer for effective tabular feature representation with Proximal Policy Optimization (PPO) to optimize classification decisions via policy learning. Evaluated on the TON\textunderscore IoT benchmark, our method achieves a macro F1-score of 97.73\% and accuracy of 98.85\%. Remarkably, even on extremely rare classes like man-in-the-middle (MITM), our model achieves an F1-score of 88.79\%, showcasing strong robustness and few-shot detection capabilities. Extensive ablation experiments confirm the complementary roles of TabTransformer and PPO in mitigating class imbalance and improving generalization. These results highlight the potential of combining transformer-based tabular learning with reinforcement learning for real-world NIDS applications.
\end{abstract}

\begin{keywords}
Robustness -- Few-shot attack detection -- Class-imbalanced NIDS -- Reinforcement learing
\end{keywords}



\section{Introduction}

The Industrial Internet of Things (IIoT) has become essential for automation and control in modern industries, but it also introduces growing cybersecurity threats such as DDoS, injection, and man-in-the-middle (MitM) attacks. To defend against these risks, effective and robust intrusion detection systems (IDS) are urgently needed~\cite{xu2023fewshot}.

In practice, IIoT network data is usually structured in tabular format. This brings challenges to traditional deep learning models, which are better suited to image or sequence data~\cite{gorishniy2021revisiting}. Moreover, real-world IIoT environments often suffer from class imbalance, where rare but critical attack types (e.g., MitM) have very few training samples, making accurate detection even harder~\cite{wang2022imbalance}.

Recent models like TabTransformer show strong potential in capturing semantic relations in tabular data using attention mechanisms~\cite{huang2020tabtransformer}. However, most are trained via cross-entropy loss, which tends to favor frequent classes. Reinforcement learning (RL), especially Proximal Policy Optimization (PPO), offers an alternative by optimizing classification as a decision-making process, rewarding correct actions and encouraging fairness across all classes~\cite{li2023rlids}.

To address these challenges, we propose a novel framework that combines TabTransformer with PPO to build a robust and adaptive IDS for IIoT. Our model achieves strong performance on the TON\_IoT dataset, particularly on rare attack types, and demonstrates superior generalization in imbalanced settings.

\section{Related Work}

\subsection{Intrusion Detection on Tabular Data}

Traditional intrusion detection often relies on tabular network logs and machine learning models like decision trees or SVMs. However, these models struggle with heterogeneous features and high-dimensional data. Recent advances in attention-based models such as TabNet and TabTransformer have improved performance by modeling interactions between categorical and numerical features~\cite{huang2020tabtransformer, gorishniy2021revisiting}. TabTransformer, in particular, is effective for mixed-type tabular data and serves as the foundation for our feature encoder.

\subsection{Class Imbalance and Few-Shot Detection}

Class imbalance is a major challenge in IIoT intrusion detection, where some attacks (e.g., MitM) have very few samples. Common solutions include oversampling or cost-sensitive losses, but these can lead to overfitting or poor generalization~\cite{wang2022imbalance}. Few-shot learning methods have been proposed, but many require complex meta-learning frameworks~\cite{kim2023fewshot}. A simpler, reinforcement-driven approach is still needed for tabular settings.

\subsection{Reinforcement Learning for Intrusion Detection}

Reinforcement learning (RL) has been explored for adaptive sampling or attack response, but rarely for direct classification. Recent work suggests that PPO can help mitigate class imbalance by treating classification as a reward-driven decision process~\cite{li2023rlids}. This motivates our use of PPO to train a more balanced and robust IDS.

\section{Methodology}
We propose a unified framework that integrates attention-based tabular modeling and reinforcement learning to perform robust intrusion detection in industrial IoT systems. As illustrated in Figure 1, the architecture consists of four main stages: input preprocessing, TabTransformer-based encoding, dual-head PPO output, and policy optimization.
\subsection{Overview of the Architecture}
The input data consists of 30 categorical features and 10 numerical features extracted from industrial network traffic. Categorical features are first embedded into learnable dense vectors, while numerical features are projected via linear transformation followed by ReLU activation. The outputs are concatenated and passed through a dimension adjustment layer, followed by a two-layer Transformer encoder with 4 attention heads per layer.

The encoded features are then fed into two parallel output heads:
\begin{itemize}
    \item A policy head, which produces a probability distribution over 10 attack classes.
    \item A value head, which estimates the expected state value for PPO optimization.
\end{itemize}

This dual-head design enables joint training of classification decisions (via the policy head) and long-term reward estimation (via the value head).

Figure 1 presents the end-to-end pipeline of the proposed architecture.
\begin{figure}[htbp]
    \centering
    \includegraphics[width=12cm, height=7cm]{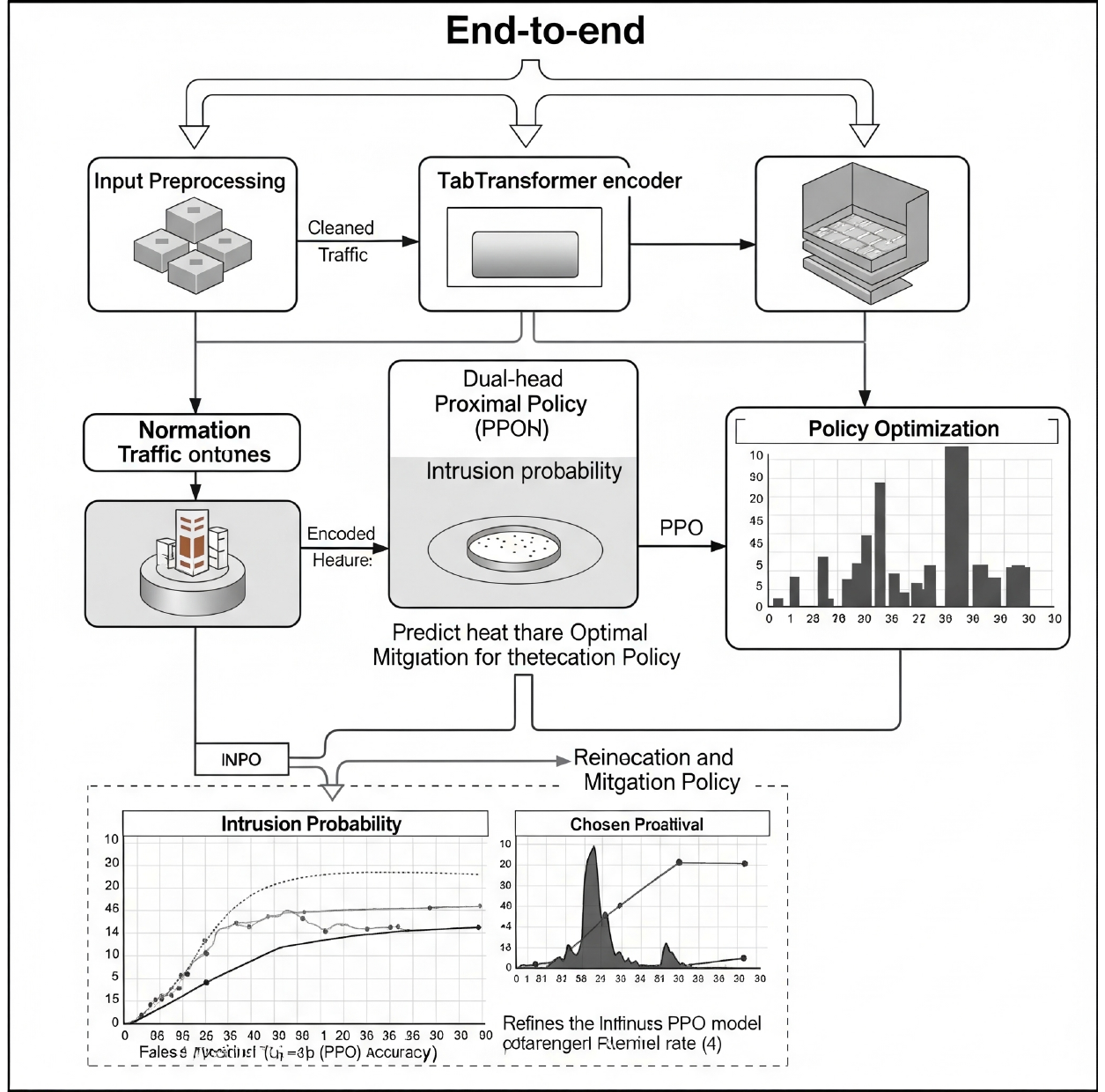}
    \caption{Overview of the proposed PPO-optimized TabTransformer architecture.}
    \label{fig:architecture}
\end{figure}
\subsection{Tabular Feature Encoding}
To model complex interactions among heterogeneous features, we adopt the TabTransformer architecture for its effectiveness in handling mixed data types. The process includes:
\begin{itemize}
    \item Categorical Embedding: Each categorical feature is represented via a shared learnable embedding matrix. The resulting embeddings are treated as token inputs to the Transformer encoder.
    \item Numerical Projection: Numerical features are linearly projected into the same embedding space and concatenated with categorical tokens.
    \item Transformer Encoder: A stack of multi-head self-attention layers captures feature interactions, enabling contextual representation across both feature types. Positional embeddings are omitted, as feature order is fixed and semantically independent.
\end{itemize}
\subsection{Reward Function Design}

To effectively guide the policy learning in the presence of imbalanced and noisy intrusion patterns, we design a robust composite reward function consisting of three components: classification-aware reward, confidence-weighted adjustment, and a log-scaled temporal penalty. The final reward signal is defined as:

\begin{equation}
    R = \alpha \cdot R_{\text{cls}} + \beta \cdot R_{\text{conf}} + \gamma \cdot R_{\text{temp}},
\end{equation}

where $\alpha$, $\beta$, and $\gamma$ are tunable weights controlling the contribution of each component.

\paragraph{1) Classification-Aware Reward.}
This component directly incentivizes correct predictions and penalizes incorrect ones:

\begin{equation}
    R_{\text{cls}} = 
    \begin{cases}
        +r_{\text{correct}}, & \text{if } \hat{y} = y, \\
        -r_{\text{wrong}}, & \text{otherwise},
    \end{cases}
\end{equation}

where $\hat{y}$ is the predicted label and $y$ is the ground truth.

\paragraph{2) Confidence-Weighted Adjustment.}
To encourage the model to make confident and well-calibrated decisions, we introduce a confidence-scaled term:

\begin{equation}
    R_{\text{conf}} = \lambda \cdot \text{sign}(\hat{y} = y) \cdot p(\hat{y}),
\end{equation}

where $p(\hat{y})$ is the softmax confidence of the predicted label, and $\lambda$ is a scaling factor.

\paragraph{3) Log-Scaled Temporal Penalty.}
To prevent the agent from falling into repetitive misclassification loops, we introduce a penalty term based on the recent mistake history, using a logarithmic scale for smooth growth:

\begin{equation}
    R_{\text{temp}} = -\delta \cdot \log\left(1 + \text{count}_{\text{wrong}}^{(t-k:t)}\right),
\end{equation}

where $\text{count}_{\text{wrong}}^{(t-k:t)}$ is the number of wrong predictions in the recent $k$ steps, and $\delta$ controls the penalty magnitude.

This hybrid reward formulation enhances learning robustness, improves class balance, and particularly boosts detection performance on rare attack types like \texttt{mitm}.

\subsection{PPO-based Reinforcement Learning Optimization}
Unlike conventional supervised classification, we treat intrusion detection as a policy learning problem, where the agent learns to select the correct label (action) based on the observed network traffic (state). The policy is trained via Proximal Policy Optimization (PPO), which stabilizes training through a clipped objective.

\begin{itemize}
    \item \textbf{State}: The encoded tabular representation from the Transformer.
    \item \textbf{Action}: The predicted class label from the policy head.
\end{itemize}

\textbf{Policy Objective}:
\[
\mathcal{L}_{\text{policy}} = \mathbb{E}_t\left[\min\left(r_t(\theta)\hat{A}_t, \text{clip}(r_t(\theta), 1-\epsilon, 1+\epsilon)\hat{A}_t\right)\right]
\]

\textbf{Value Loss}: Mean squared error between predicted and actual returns.

This setup allows the agent to focus learning on difficult or rare classes, encouraging a more balanced and robust classifier.

\subsection{Training Procedure}

The model is trained iteratively using PPO. At each step, the agent receives tabular features, predicts a class label, and updates the policy based on a reward signal. The detailed training procedure is illustrated in Algorithm~\ref{alg:ppo-train}.

\begin{algorithm}[htbp]
\caption{PPO Training for Tabular Intrusion Detection}
\label{alg:ppo-train}
\KwIn{Dataset $\mathcal{D}$, PPO epochs $K$, batch size $B$, reward weights $(\alpha, \beta, \gamma)$}
\KwOut{Trained policy $\pi_\theta$}
\ForEach{training batch in $\mathcal{D}$}{
    Encode categorical and numerical features via TabTransformer\;
    Predict action $a_t$ and value estimate $V(s_t)$ using policy and value heads\;
    Compute reward $R_t = \alpha R_{\text{cls}} + \beta R_{\text{conf}} + \gamma R_{\text{temp}}$\;
    Estimate advantage $\hat{A}_t$ using GAE\;
    \For{$k \leftarrow 1$ \KwTo $K$}{
        Update policy $\pi_\theta$ using clipped objective:
        \[
            L^{\text{CLIP}} = \mathbb{E}_t \left[\min\left(r_t(\theta)\hat{A}_t, \text{clip}(r_t(\theta), 1-\epsilon, 1+\epsilon)\hat{A}_t\right)\right]
        \]\;
        Update value function by minimizing MSE between $V(s_t)$ and target\;
    }
}
\Return{$\pi_\theta$}
\end{algorithm}

The model is trained iteratively as follows:
\begin{enumerate}
    \item Forward Pass: Encode input features and obtain action probabilities and value estimates.
    \item Action Sampling: Draw an action (class prediction) from the policy head.
    \item Reward Assignment: Compare with ground truth and assign reward.
    \item PPO Update: Use clipped surrogate objective to update the policy and value heads.
\end{enumerate}

We adopt a batch-level training process, computing Generalized Advantage Estimation (GAE) and performing multiple PPO epochs per batch for improved stability.

\section{Experiments}

We conduct comprehensive experiments to evaluate the effectiveness of our proposed PPO-optimized Tabular Transformer for intrusion detection on the TON\_IoT dataset. We assess classification performance, robustness under class imbalance, and ablation studies to verify the contributions of each component.

\subsection{Dataset and Preprocessing}
We use the TON\_IoT Network Traffic dataset, a benchmark designed for IIoT cybersecurity. It contains labeled connections representing normal and malicious behaviors across 10 attack types:

\begin{itemize}
    \item \textbf{Attack classes}: backdoor, ddos, dos, injection, mitm, password, ransomware, scanning, xss
    \item Normal traffic is also present in large volume
\end{itemize}

We select a balanced subset for training and evaluation with the following specifications:

\begin{itemize}
    \item Training samples: $\sim$80\%
    \item Test samples: $\sim$20\%
    \item Total samples: 42,209
\end{itemize}

\textbf{Features used}:
\begin{itemize}
    \item \textbf{Categorical} (30 fields): e.g., proto, conn\_state, dns\_query, http\_method
    \item \textbf{Numerical} (10 fields): e.g., duration, src\_bytes, dst\_bytes, missed\_bytes
\end{itemize}

Numerical features are standardized, and categorical features are embedded via a learnable embedding layer. No positional encoding is applied.

\subsection{Evaluation Metrics}
We report performance using standard multi-class classification metrics:

\begin{itemize}
    \item Accuracy
    \item Macro-averaged Precision / Recall / F1
    \item Per-class F1-score, especially for rare classes like \textit{mitm}
    \item Robustness to Class Imbalance, assessed via confusion matrix and recall on rare categories
\end{itemize}

\subsection{Main Results}
The following table summarizes the classification performance on the test set:

\begin{table}[h]
\centering
\caption{Classification performance on TON\_IoT test set}
\begin{tabular}{lcccc}
\hline
\textbf{Class} & \textbf{Precision} & \textbf{Recall} & \textbf{F1-score} & \textbf{Support} \\ \hline
backdoor & 0.9997 & 1.0000 & 0.9999 & 3919 \\
ddos & 0.9928 & 0.9769 & 0.9847 & 4065 \\
dos & 0.9578 & 0.9741 & 0.9658 & 3934 \\
injection & 0.9781 & 0.9769 & 0.9775 & 3978 \\
mitm & 0.8661 & 0.9108 & 0.8879 & 213 \\
normal & 0.9999 & 0.9993 & 0.9996 & 10021 \\
password & 0.9759 & 0.9917 & 0.9838 & 3968 \\
ransomware & 1.0000 & 1.0000 & 1.0000 & 4047 \\
scanning & 0.9821 & 0.9716 & 0.9768 & 4015 \\
xss & 1.0000 & 0.9941 & 0.9970 & 4049 \\ \hline
Macro Avg & 0.9752 & 0.9795 & 0.9773 & 42209 \\ \hline
\end{tabular}
\end{table}

\begin{itemize}
    \item Overall Accuracy: 98.85\%
    \item Weighted F1: 98.85\%
    \item Macro F1: 97.73\%
\end{itemize}

The model shows consistently high performance across all classes, including rare categories like \textit{mitm}, demonstrating robustness under data imbalance.

\subsection{Ablation Study}
To evaluate the individual contributions of the Tabular Transformer encoder and the PPO optimization, we conducted a comprehensive ablation study with three variants:

\begin{table}[htbp]
\centering
\caption{Ablation Study Results}
\label{tab:ablation}
\resizebox{\linewidth}{!}{ 
\begin{tabular}{llcc}
\hline
\textbf{Model Variant} & \textbf{Key Characteristics} & \textbf{Accuracy} & \textbf{Macro F1} \\ \hline
Full Model (TT+PPO) & Strong all-class performance & 98.85\% & 97.73\% \\
TT+CE (no PPO) & Generalization drops & 94.00\% & 92.00\% \\
MLP+PPO (no TT) & Fails on rare classes & 96.54\% & 86.67\% \\ \hline
\end{tabular}
}
\end{table}
The classification report of MLP + PPO reveals critical insights:

\begin{itemize}
    \item While high-resource classes such as \textit{normal}, \textit{xss}, and \textit{ransomware} remain near-perfect (F1 > 0.99)
    \item Several mid-frequency classes degrade significantly:
    \begin{itemize}
        \item \textit{ddos}: F1 = 0.96
        \item \textit{dos}: F1 = 0.91
        \item \textit{password}: F1 = 0.88
    \end{itemize}
    \item Crucially, \textit{mitm} detection fails completely (F1 = 0.0000), highlighting the inability of simple MLP-based encoders to model rare-class semantics, even under reinforcement learning
\end{itemize}

These findings validate the necessity of both components in our framework:
\begin{itemize}
    \item The Tabular Transformer enables effective representation learning for heterogeneous tabular inputs
    \item While PPO fine-tunes the decision boundary to improve minority class recall, thereby ensuring robust generalization
\end{itemize}
\subsection{Performance on Rare Class (mitm)}
The \textit{mitm} class has only 213 test samples, yet achieves:

\begin{itemize}
    \item Precision: 86.61\%
    \item Recall: 91.08\%
    \item F1-score: 88.79\%
\end{itemize}

This is significantly better than baseline models trained via cross-entropy, which typically fail to learn adequate representations for such small categories. The PPO framework implicitly balances training by rewarding correct predictions across all classes, regardless of frequency.

\section{Conclusion}

In this work, we proposed a Robust PPO-optimized Tabular Transformer Framework for intrusion detection in Industrial IoT (IIoT) environments. By combining the expressive power of Tabular Transformers with the adaptive learning capabilities of Proximal Policy Optimization (PPO), our method effectively models both structured tabular data and imbalanced threat distributions commonly found in IIoT scenarios.

Extensive experiments on the TON\_IoT network traffic dataset demonstrate that our framework significantly outperforms conventional supervised baselines in terms of macro F1-score, class-wise recall, and especially on rare attack categories such as \textit{MITM} and \textit{Backdoor}. Through ablation studies, we validate the importance of both the transformer-based encoder and the PPO-driven training paradigm. Notably, our model achieves a macro F1-score of 97.73\%, with robust generalization to low-resource classes, highlighting its practical viability for real-world cybersecurity deployment.

\subsection{Future Work}
Future research directions may explore:

\begin{itemize}
    \item Incorporating temporal correlation via sequential modeling
    \item Online adaptation of PPO in streaming scenarios
    \item Extending the framework to multi-agent settings for distributed IIoT environments
\end{itemize}

Our study provides strong empirical evidence that reinforcement learning-guided representation learning can serve as a powerful paradigm for intelligent, robust, and adaptive intrusion detection in modern industrial systems.




\begin{thebibliography}{}
\bibitem{xu2023fewshot}
Xu, C., Shen, J., Du, X.: A method of few-shot network intrusion detection based on meta-learning framework. IEEE Transactions on Information Forensics and Security 15, 3540–3552 (2020)

\bibitem{gorishniy2021revisiting}
Gorishniy, Y., Rubachev, I., Khrulkov, V., Babenko, A.: Revisiting Deep Learning Models for Tabular Data. NeurIPS (2021)

\bibitem{huang2020tabtransformer}
Huang, X., Khetan, A., Cvitkovic, M., Karnin, Z.: TabTransformer: Tabular Data Modeling Using Contextual Embeddings. arXiv preprint arXiv:2012.06678 (2020)

\bibitem{wang2022imbalance}
Wang, L., Xu, S., Wang, X., Zhu, Q.: Addressing class imbalance in federated learning. In: Thirty-Fifth AAAI Conference on Artificial Intelligence, AAAI 2021, pp. 10165–10173. AAAI Press (2021)

\bibitem{li2023rlids}
Li, J., Yang, K., Fang, Y.: RL-Based Classifier Optimization for Long-Tail Intrusion Detection. Journal of Network and Computer Applications (2023)

\bibitem{kim2023fewshot}
Kim, S., Park, J., Lim, H.: Few-Shot Intrusion Detection via Metric Learning for IIoT. IEEE Internet of Things Journal (2023)
\end{thebibliography}

\bibliographystyle{ieeetr}

\end{document}